\documentclass[runningheads]{llncs}
\usepackage[T1]{fontenc}
\usepackage{graphicx,verbatim}
\usepackage{hyperref}
\usepackage{multirow}
\usepackage{caption} 
\usepackage{amsmath,amssymb} 
\usepackage{graphicx}      
\usepackage{booktabs}  
\usepackage{xcolor}

\begin{document}
\title{Self-supervised Disentanglement of Disease Effects from Aging in 3D Medical Shapes}
\titlerunning{Self-supervised Disentanglement of Disease Effects in 3D Medical Shapes}
%

\author{Jakaria Rabbi\inst{1}, Nilanjan Ray\inst{1}, Dana Cobzas\inst{2}}  
\authorrunning{Rabbi et al.}
\institute{$^{1}$Department of Computing Science, University of Alberta, Edmonton, Canada\\$^{2}$Department of Computer Science, MacEwan University, Edmonton, Canada\\
    \email{jakaria@ualberta.ca, nray1@ualberta.ca, cobzasd@macewan.ca}}
  
\maketitle              
\begin{abstract}
Disentangling pathological changes from physiological aging in 3D medical shapes is crucial for developing interpretable biomarkers and patient stratification. However, this separation is challenging when diagnosis labels are limited or unavailable, since disease and aging often produce overlapping effects on shape changes, obscuring clinically relevant shape patterns. To address this challenge, we propose a two-stage framework combining unsupervised disease discovery with self-supervised disentanglement of implicit shape representations. In the first stage, we train an implicit neural model with signed distance functions to learn stable shape embeddings. We then apply clustering on the shape latent space, which yields pseudo disease labels without using ground-truth diagnosis during discovery. In the second stage, we disentangle factors in a compact variational space using pseudo disease labels discovered in the first stage and the ground truth age labels available for all subjects. We enforce separation and controllability with a multi-objective disentanglement loss combining covariance and a supervised contrastive loss. On ADNI hippocampus and OAI distal femur shapes, we achieve near-supervised performance, improving disentanglement and reconstruction over state-of-the-art unsupervised baselines, while enabling high-fidelity reconstruction, controllable synthesis, and factor-based explainability. Code and checkpoints are available at \href{https://github.com/anonymous-submission01/medical-shape-disentanglement }{\textcolor{blue}{GitHub}}.

\keywords{Disentanglement \and Implicit Neural Representation  \and Interpretability \and Variational Autoencoder \and Self-supervised machine learning}

\end{abstract}

\section{Introduction}
Evaluating morphological variation in anatomical shapes is a crucial objective in computational medicine, where subtle changes can serve as imaging biomarkers for diagnosis, prognosis, and patient stratification \cite{cerrolaza2019computational,franko2013evaluating}. In neurodegeneration, hippocampal atrophy patterns are widely studied in cohorts such as the Alzheimer's Disease Neuroimaging Initiative (ADNI) \cite{Petersen2010ADNI}. 
Disease-related effects are strongly entangled by physiological aging, as they may produce overlapping shape changes, making it challenging to isolate clinically meaningful disease-associated variations, especially when diagnosis labels are limited, noisy, or unavailable \cite{gerardin2009multidimensional}. Moreover, separating disease effects from healthy controls is also crucial in diseases like osteoarthritis, where aging effects are less pronounced \cite{ambellan2019automated}. 

\begin{figure}[t]
  \centering
  \includegraphics[width=11cm]{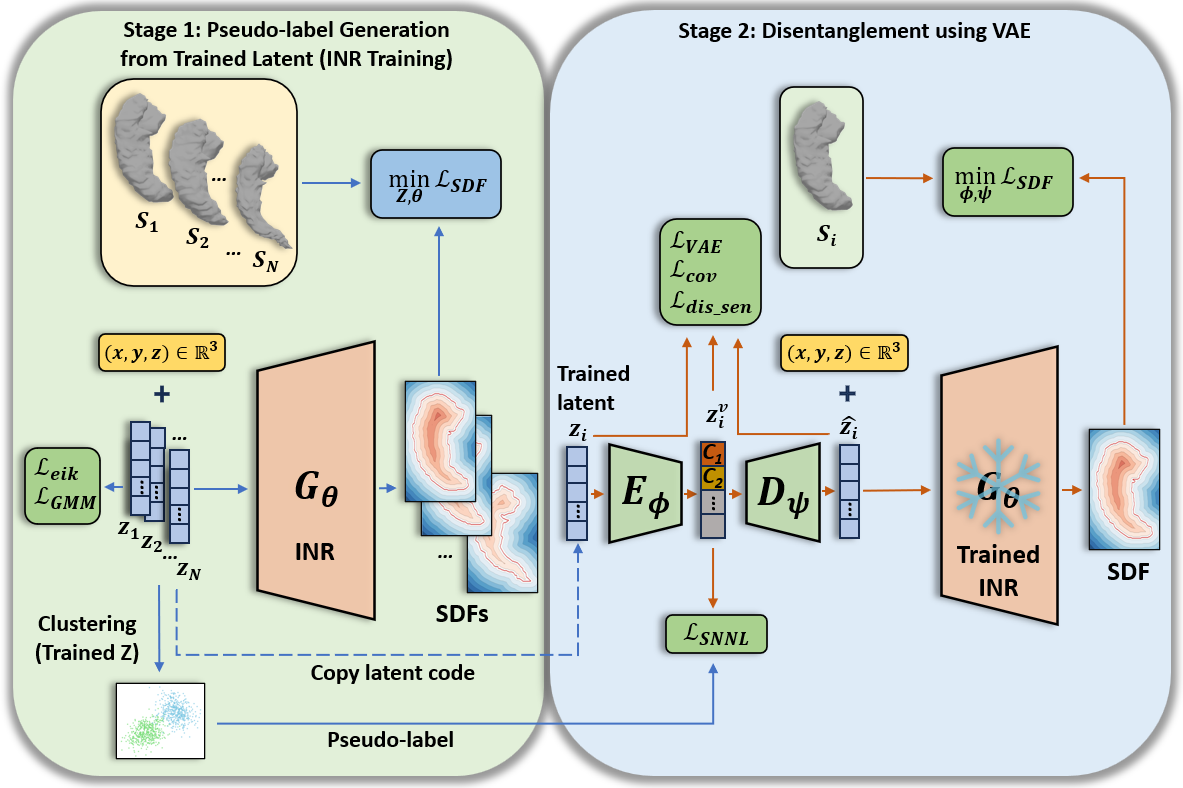}
  \caption{Stage 1: Learns per-shape codes ($\mathbf{z}_i$) and a INR-nased SDF decoder ($G_{\theta}$). Unsupervised clustering is applied on the learned codes to create pseudo-labels.
  Stage 2: A variational autoencoder (VAE) models the distribution of shape codes and learns latents ($\mathbf{z}^v_i$).
  Pseudo-labels and age labels are used for disentangling specific latent variables, while frozen $G_{\theta}$ is used for reconstruction. }
  \label{fig:pipeline}
\end{figure}
Representation learning has shifted medical shape modeling from hand-crafted descriptors and statistical models toward deep generative approaches \cite{molaei2023implicit,dannecker2024cina}. Implicit neural representations (INR) based on signed distance functions (SDF) provide high-fidelity continuous geometry and have enabled strong reconstruction, interpolation, and completion performance \cite{amiranashvili2024learning,park2019deepsdf,Mescheder2019Occupancy}, but most implicit models focus on reconstruction quality and do not directly address factor separation or label scarcity. In parallel, disentanglement in generative models has been extensively studied using variational objectives and independence-promoting regularizers \cite{burgess2018understanding,chen2018isolating,kumar2017variational}, yet purely unsupervised disentanglement is fundamentally ambiguous without additional inductive biases or supervision \cite{Locatello2019Challenging}. Within medical imaging, weak or structured supervision has been used to disentangle disease \cite{zhang2025surface,Ouyang2022AgingDisease}, and supervised shape disentanglement has been explored when labels are available \cite{Kiechle2023ISBI,Rabbi2024MELBA}. Still, there remains a gap for shape frameworks that simultaneously provide unsupervised/self-supervised disease discovery and disentanglement within a single pipeline compatible with implicit representations.\par
To address these challenges, we introduce a two-stage framework that couples unsupervised disease discovery with self-supervised disentanglement in implicit shape representations. In the first stage, we train an INR model to learn stable latent shape embeddings and encourage a clustering organization of the shape latent space through a Gaussian mixture prior, yielding pseudo disease labels. In the second stage, we map learned shape codes into a compact variational latent space and disentangle disease and age using pseudo disease supervision and continuous age labels, with a multi-objective loss \cite{Rabbi2024MELBA}. We validate our model on hippocampus meshes from ADNI \cite{Petersen2010ADNI} and OAI \cite{ambellan2019automated}, showing that pseudo-labels improve disentanglement in unlabeled and low-label settings and support interpretable traversals that control disease and age-associated morphology.

\par\textbf{Contributions:} \textit{(i)} We present the first INR-based self-supervised disease disentanglement framework for 3D medical shapes. \textit{(ii)} We disentangle disease and age using a multi-objective disentanglement loss. (\textit{iii}) On ADNI hippocampus shapes, our method outperforms unsupervised baselines and also generalizes well to an osteoarthritis (OAI) dataset.

\section{Method}

\label{sec:method}

\subsection{Overview and Problem Setup}
We propose a two-stage pipeline (Fig.~\ref{fig:pipeline}) that first learns implicit shape codes with an INR to discover healthy and diseased clusters, then trains a pseudo and age label-guided VAE on these codes while freezing the INR as a shape renderer to preserve output shape consistency from reconstructed shape codes.

We represent each surface $\mathcal{S}_i$ by samples of a signed distance field.
For shape $i$, let
$\Omega_i=\{(\mathbf{p}_{ij}, s_{ij})\}_{j=1}^{m}$ denote $m$ query points
$\mathbf{p}_{ij}\in\mathbb{R}^3$ with ground-truth signed distances $s_{ij}\in\mathbb{R}$.
Stage-1 learns $d$ dimensional codes, $\mathbf{z}_i\in\mathbb{R}^{d}$ using an INR model
\cite{park2019deepsdf}. Stage-2 learns a $k$ ($k<<d$) dimensional latents
$\mathbf{z}_i^v\in\mathbb{R}^{k}$ over the shape codes using a VAE.

\subsection{Stage-1: Implicit Representation Learning}
We represent each shape as the zero level set of a continuous SDF.
Let $G_{\theta}:\mathbb{R}^3 \times \mathbb{R}^{d} \rightarrow \mathbb{R}$ be an implicit decoder that predicts
the signed distance at a query point $\mathbf{p}\in\mathbb{R}^3$ conditioned on a shape code
$\mathbf{z}_i\in\mathbb{R}^{d}$. Given SDF samples
$\Omega_i$ for shape $i$, we learn $\theta$ and per-shape codes by minimizing $\mathcal{L}_{\text{INR}}(\mathbf{Z};\theta)$, $\mathbf{Z} \equiv \{\mathbf{z}_i\}_{i=1}^N$:
\begin{equation}
\begin{split}
\mathcal{L}_{\text{INR}}(\mathbf{Z},\theta)
&=
\underbrace{\mathcal{L}_{\text{SDF}}(\mathbf{Z},\theta)}_{\text{SDF Loss}}
\;+\;
\lambda_{\text{eik}}\,
\underbrace{\mathcal{L}_{\text{eik}}(\mathbf{Z},\theta)}_{\text{eikonal}}
\;+\;
\lambda_{\text{gmm}}\,
\underbrace{\mathcal{L}_{\text{GMM}}(\mathbf{Z})}_{\text{mixture prior}}.
\end{split}
\label{eq:sdf_split_gmm}
\end{equation}
Following DeepSDF \cite{park2019deepsdf}, we use the same architecture for our INR model and fit the SDF values by a pointwise distance penalty for reconstruction:
\begin{align}
\mathcal{L}_{\text{SDF}}(\mathbf{Z};\theta)
&=
\frac{1}{|\Omega|}\sum_{(\mathbf{p},s)\in\Omega}
\rho\!\left(G_{\theta}(\mathbf{p},\mathbf{Z}) - s\right)\;+\;
\lambda_{\text{reg}}\,
\mathcal{L}_{\text{SDF-reg}}(\mathbf{Z}),
\label{eq:sdf_rec}
\end{align}
where $\rho(\cdot)$ is a simple per-sample loss. We keep a small $\ell_2$ penalty, $\mathcal{L}_{\text{SDF-reg}}(\mathbf{Z})$ on codes to prevent unbounded growth and to stabilize optimization. We also add Eikonal \cite{gropp2020igr} loss that encourages the decoder to act like a distance field:
\begin{align}
\mathcal{L}_{\text{eik}}(\mathbf{Z};\theta)
&=
\frac{1}{|\Omega|}\sum_{\mathbf{p}\in\Omega}
\left(\|\nabla_{\mathbf{p}} G_{\theta}(\mathbf{p},\mathbf{Z})\|_2 - 1\right)^2.
\label{eq:eikonal}
\end{align}
Stage-1 shape codes exhibit a bimodal structure; therefore, we make this separation stronger and improve pseudo-label purity by introducing a
two-component Gaussian mixture prior (GMM) over the codes:
\begin{equation}
\mathcal{L}_{\text{GMM}}(\mathbf{Z})
= -\log \left(\sum_{m=1}^{2} \pi_m\,
\mathcal{N}\!\left(\mathbf{Z}\mid \boldsymbol{\mu}_m,\mathrm{diag}(\boldsymbol{\sigma}_m^2)\right)\right).
\label{eq:gmm_prior}
\end{equation}
Here, $\pi_m$ are mixture weights ($\sum_m \pi_m=1$), and each component $m$ has a learnable mean
$\boldsymbol{\mu}_m$ and diagonal covariance $\mathrm{diag}(\boldsymbol{\sigma}_m^2)$.
Minimizing $\mathcal{L}_{\text{GMM}}$ increases the likelihood of codes aligning with different clusters.
\newline
\textbf{Pseudo-label Discovery in Shape Code:} We fit a 2-component GMM on the learned shape codes ($\mathbf{z_i}$) and assign $\tilde{y}_i=\arg\max_m p(m\mid \mathbf{z_i})$.

\subsection{Stage-2: Self-supervised Disentanglement and INR}
Stage-2 models the distribution of stage-1 shape codes with a VAE, learning a low-dimensional latent code. The encoder $E_{\phi}$ outputs $\mathbf{Z}^v \equiv \{\mathbf{z}^v_i\}_{i=1}^N$, and the decoder $D_{\psi}$ reconstructs the shape code as $\hat{\mathbf{Z}}\equiv \{\hat{\mathbf{z}}_i\}_{i=1}^N$. We use residual MLP stacks for both encoder and decoder. Encoder contains hidden widths 256 and 128 that maps the input to VAE latent code. The detector consisting of hidden widths 128, 256, and 256 maps the latent code to the reconstructed input. Each stage contains one residual block with GeLU activations and LayerNorm. We pass $\hat{\mathbf{Z}}$ through the pretrained implicit decoder $G_{\theta}$ with $\theta$ fixed.
\newline
\textbf{Training objective:}
\label{sec:stage2_objective}
Stage-2 optimizes a weighted sum of five terms:
\begin{align}
\label{eq:stage2_total}
\mathcal{L}_{\text{disentangle}}(\phi,\psi)
&=
\underbrace{\mathcal{L}_{\text{VAE}}\!\big(q_{\phi}(\mathbf{Z}^v\mid \mathbf{Z}),\hat{\mathbf{Z}}\big)}_{\text{VAE loss}}
+\lambda_{\text{snnl}}\,\underbrace{\mathcal{L}_{\text{SNNL}}(\mathbf{Z}^v,\tilde{y},c)}_{\text{supervised loss}}
+\lambda_{\text{cov}}\,\underbrace{\mathcal{L}_{\text{cov}}(\mathbf{Z}^v)}_{\text{decorrelation}}
\nonumber\\[-1pt]
&\quad
+\lambda_{\text{dis\_sen}}\,\underbrace{\mathcal{L}_{\text{dis\_sen}}(\mathbf{Z}^v,\psi,c)}_{\text{dist. \& sens.}}
+\lambda_{\text{SDF}}\,\underbrace{\mathcal{L}_{\text{SDF}}(\hat{\mathbf{Z}})}_{\text{SDF loss.}} .
\end{align}
Here, $c$ are the designated coordinates for supervised disentanglement. We group code reconstruction and KL regularization into a single term,
$\mathcal{L}_{\text{VAE}}\!\big(q_{\phi}(\mathbf{Z}^v\mid \mathbf{Z}),\hat{\mathbf{Z}}\big)$ consisting of reconstruction loss, $\mathcal{L}_{\text{code}}(\mathbf{Z},\hat{\mathbf{Z}})$ and KL loss, $\mathcal{L}_{\text{KL}}\!\big(q_{\phi}(\mathbf{Z}^v\mid \mathbf{Z})\big)$.
The reconstruction part recovers the shape code and is multiplied by $\lambda_{code}$,
while the KL term regularizes the posterior toward the unit Gaussian prior, multiplied by $\beta$.
We apply a soft nearest-neighbor loss (SNNL) \cite{Rabbi2024MELBA} on $Z^{v}_c$:
\begin{equation}
\label{eq:snnl_cls_compact}
\mathcal{L}_{\mathrm{SNNL}}(\mathbf{Z}_v,\tilde{y};c,Th)
=\sum_c
-\frac{1}{b}\sum_{i=1}^{b}\log\!\left(
\frac{\sum_{j\in\mathcal{P}_i(Th)}\alpha^{c}_{ij}}
{\lambda_1\sum_{k\neq i}\alpha^{c}_{ik}\;+\;\lambda_2\sum_{j\in\mathcal{P}_i(Th)}\alpha^{\neg c}_{ij}}
\right).
\end{equation}
Here, $\tilde{y}$ is the label (for classification threshold $Th=0$), $\mathcal{P}_i(Th)=\{j\neq i:\,|\tilde{y}_j-\tilde{y}_i|\le Th\}$.
We define the affinity
$\alpha^{c}_{ij}=\exp\!\big(-\|z_{i,c}^{\,v}-z_{j,c}^{\,v}\|^2/T\big)$ and
$\alpha^{\neg c}_{ij}=\exp\!\Big(-[1/(|\mathcal{D}_{\neg t}|\,T)]\sum_{d\in\mathcal{D}_{\neg t}}\|z_{i,d}^{\,v}-z_{j,d}^{\,v}\|^2\Big)$,
where $T>0$ is an adaptive temperature estimated from batch distances (low variance $\rightarrow$ small $T$) and $\mathcal{D}_{\neg t}=\{1,\dots,k\}\setminus\{t\}$. The weights $\lambda_1,\lambda_2\ge 0$ balance (i) repulsion among all samples in the target coordinate
and (ii) discouraging pseudo-label similarity leaking into non-target coordinates. We also penalize the off-diagonal entries (equation \ref{eq:l_cov}) for disentanglement where $\mathbf{C_v}=\mathrm{Cov}(\mathbf{Z}^v)$ is the batch covariance of $\mathbf{Z}^v$.
\begin{equation}
\label{eq:l_cov}
\mathcal{L}_{\text{cov}}(\mathbf{Z}^v)
=
\left\|\mathbf{C_v}-\mathrm{diag}(\mathbf{C_v})\right\|_F^2 .
\end{equation}
When SNNL is applied for disease disentanglement, we empirically observed two failure modes: variance collapse of $\mathbf{Z}^v_c$ toward $0$ \cite{bardes2022vicreg}, and coordinate inactivity, where changing
$\mathbf{Z}^v_c$ yields negligible change in the reconstructed code \cite{lucas2019posteriorcollapse}). We therefore add a compact regularizer on the
designated coordinate in equation \ref{eq:l_zt} inspired by isometric representation learning. \cite{lee2022regularized}.
\begin{equation}
\label{eq:l_zt}
\mathcal{L}_{dis\_sen} \;=\; (s_c-\bar{s}_{\neg c})^2 \;+\; \Big(\max(0,\eta-\alpha_c)/\eta\Big)^2 .
\end{equation}
Here $s_c=\mathrm{Std}_{\mathcal{B}}(\mathbf{Z}^v_c)$, $\bar{s}_{\neg c}=\frac{1}{k-1}\sum_{d\neq c}\mathrm{Std}_{\mathcal{B}}(\mathbf{Z}^v_d)$,
and $\alpha_c=\frac{1}{b}\sum_{i=1}^{b}\big\|D_{\psi}(\mathbf{z}^v_{\,i}+\varepsilon\mathbf{e}_c)-D_{\psi}(\mathbf{z}^v_{\,i}-\varepsilon\mathbf{e}_c)\big\|_2$
is the average finite-difference effect of perturbing only $Z^{v}_c$ (with $\mathbf{e}_c$ the $c$-th basis vector). We reuse one of the INR losses ( $\mathcal{L}_{\text{shape}}(\hat{\mathbf{Z}};G_{\theta})$) from stage-1 on $\hat{\mathbf{Z}}$ with $\theta$ fixed.

\section{Experiments}
\label{sec:experiments}

\subsection{Datasets}
We use two datasets: hippocampal shapes from the ADNI-1 3T MRI dataset \cite{Petersen2010ADNI} (1~mm isotropic), including left/right segmentation masks with ground-truth volumes, and distal femur shapes from the Osteoarthritis Initiative (OAI) dataset (with femur segmentation) \cite{ambellan2019automated} for cross-disease evaluation. For ADNI, we generate 3D meshes using marching cubes \cite{lorensen1998marching} followed by smoothing. We validate 819 ADNI (AD and cognitive normal - CN) and 191 (healthy and 
OA) OAI scans with patient-disjoint, stratified splits of 80\%/10\%/10\% (train/val/test), preserving age, sex, and diagnosis ratios. All scans are used for representation learning, and unsupervised/self-supervised disentanglement following Horan et al. \cite{horan2021when}, while the splits are used for supervised ablation.

\subsection{Implementation Details and Metrics}
\label{sec:impl_metrics}
We train a two-stage pipeline on three NVIDIA Titan RTX (24~GB) GPUs. Stage-1 is trained for 2000 epochs on SDF samples (16 shapes/batch, 16{,}384 samples/shape; clamping=0.1; grad-clip=1.0) with $\lambda_{eik}=\lambda_{reg}=10^{-4}$, $\lambda_{gmm}=10^{-3}$ ($K{=}2$), and step LR decay (lr=0.001). Stage-2 trains a VAE with the shape codes as input/output while keeping the SDF decoder frozen as a renderer; it runs for 2000 epochs (32 shapes/batch, 16{,}384 samples/shape) with $\lambda_{reg}=10^{-4}$, $\lambda_{code}=0.44$, $\beta=0.008$, $\lambda_{snnl}=0.77$, $(\lambda_1=\lambda_2=0.5)$, 
$Th=.05$,
$\lambda_{cov}=0.007$, and $\lambda_{dis\_sen}=0.56$ ($\varepsilon=0.02$, $\eta=0.02$). 

We find the best parameters by using the Optuna framework \cite{akiba2019optuna} and for Stage-2 ablations, run each setting with three random seeds, reporting average performance. We evaluate disentanglement using SAP \cite{Locatello2019Challenging} and correlations between latent dimensions and covariates (age, diagnosis), with diagnosis classification, age regression, and shape reconstruction. All methods use the latent codes from our stage-1 because most unsupervised disentanglement methods perform poorly on the shape space. We report all comparisons and ablations on ADNI, and include OAI results for our method only in the last row of Table~\ref{tab:quant_main} due to space constraints.

\begin{figure}[t]
\centering
\begin{minipage}[t]{0.45\linewidth}
\centering
\null
\captionof{table}{Stage-1 ablation:
purity (Pur (\%)$\uparrow$), and mean-volume gap ($\Delta\overline{V}\uparrow$, mm$^3$) for shapes between the two clusters. Average volume of AD is lower compared to CN which is clearly represented by stage-1 clusters.}
\label{tab:stage1_ablation}
\setlength{\tabcolsep}{2.5pt}
\renewcommand{\arraystretch}{0.95}
\scriptsize
\begin{tabular}{lccc}
\hline
Losses & Pur (\%)$\uparrow$  & $\Delta\overline{V} (mm^3)\uparrow$ & Recon. $\downarrow$ \\
\hline
Rec + $\ell_2$ & 72.43 &  1154 &  0.0014\\
+ Eikonal      & 80.19 &  1317 &  \textbf{0.0013}\\
+ GMM prior    & \textbf{82.37} &  \textbf{1401} & 0.0015  \\
\hline
\end{tabular}
\end{minipage}\hfill
\begin{minipage}[t]{0.45\linewidth}
\centering
\null
\includegraphics[width=4.0cm]{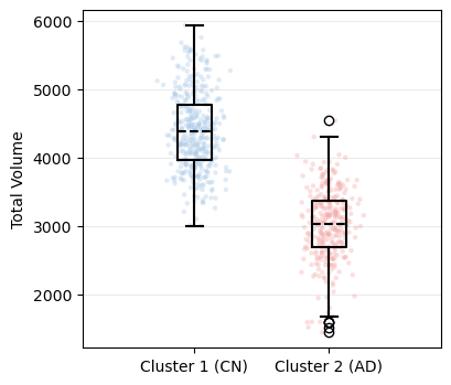}

\caption{Volume distribution of shapes in stage-1 cluslters.}
\label{fig:stage1_ablation_vis}
\end{minipage}
\end{figure}

\section{Results}
\subsection{Clustering quality, Volume analysis and Disentanglement}
We evaluate \textbf{stage-1} INR training in Table~\ref{tab:stage1_ablation}. Cluster purity (\textbf{Pur}) is computed with AD/CN labels only for evaluation by averaging the majority-class fraction per cluster. Adding Eikonal improves Pur (72.43$\rightarrow$80.19) and reconstruction (0.0014$\rightarrow$0.0013), indicating smoother SDF geometry. Adding a GMM prior gives the best purity (82.37) and largest mean volume difference of shapes ($\boldsymbol{\Delta\overline{V}}=1401~mm^3$) between two clusters with a small reconstruction error using the chamfer distance (CD) increase (0.0015). Fig.~\ref{fig:stage1_ablation_vis} shows that the mean volumes differ and the distributions overlap, implying clustering is not based on volume. Cluster 1 represents CN and cluster 2 represents AD, as CN cohorts have higher hippocampal volumes \cite{chupin2009fully}.  Similar process is followed for OAI dataset. Shape-space clustering and HLLE+ICA \cite{horan2021when}, does not produce meaningful disentanglement; therefore, we adopt our Stage-1 method.

\begin{table}[t]
\centering
\caption{Comparison of representation methods, unsupervised disentanglement models, and our method.
$\uparrow$ / $\downarrow$ indicates higher / lower is better, respectively. All rows except the last one show the results of the ADNI dataset. The last row shows the results from the OAI dataset with the percentage increase or decrease from the second-best or best method.}
\label{tab:quant_main}
\setlength{\tabcolsep}{1.5pt}
\renewcommand{\arraystretch}{1.0}
\scriptsize
\begin{tabular*}{\linewidth}{@{\extracolsep{\fill}}lccccccc}
\hline
\multirow{2}{*}{Method}
& \multicolumn{3}{c}{Disease} & \multicolumn{3}{c}{Age} & \multirow{2}{*}{Recon.$\downarrow$} \\
\cline{2-4}\cline{5-7}
& \rule{0pt}{2.6ex}SAP $\uparrow$ & Correlation$\uparrow$ & Accuracy$\uparrow$
& \rule{0pt}{2.6ex}SAP$\uparrow$ & Correlation$\uparrow$ & RMSE$\downarrow$
& \\
\hline
PCA       & 0.14 & 0.57 & 67.38 & 0.03 & 0.22 & 0.162 & 0.0020 \\
ICA       & 0.14 & 0.61 & 63.15 & 0.02& 0.20 & 0.162 & 0.0021 \\
HLLE+ICA  & 0.18 & 0.67 & 71.76 & 0.02 & 0.18 & 0.161 & N/A \\
\hline
$\beta$-VAE    & 0.15 & 0.47 & 51.27 & 0.61 & 0.79 & 0.065 & \textbf{0.0016} \\
$\beta$-TCVAE  & 0.16 & 0.49 & 51.94 & 0.64 & 0.82 & 0.068 & 0.0017 \\
DIP-VAE        & 0.18 & 0.48 & 53.61 & 0.63 & \textbf{0.86} & 0.062 & 0.0017 \\
\hline
w/ fixed T & 0.30 & 0.69 & 76.31 & 0.63 & 0.82 & 0.065 & 0.0018 \\
w/o cov    & 0.34 & \textbf{0.73} &  76.55 & 0.63 & 0.83 & 0.068 & 0.0018 \\
\textbf{Ours}  & \textbf{0.38} & \textbf{0.73} & \textbf{78.67} & \textbf{0.67} & \textbf{0.86} & \textbf{0.061} & 0.0019 \\ 
\hline
\noalign{\vskip 1pt}
\textbf{\shortstack{Ours\\(OAI )}} 
& \textbf{\shortstack{0.25\\(+32\%)}} 
& \textbf{\shortstack{0.55\\(+1\%)}} 
& \textbf{\shortstack{63.67\\(+2\%)}}
& \textbf{\shortstack{0.41\\(+5\%)}} 
& \textbf{\shortstack{0.52\\(+7\%)}} 
& \textbf{\shortstack{0.091\\(+5\%)}} 
& \shortstack{0.0016\\(-10\%)} \\
\hline
\end{tabular*}
\end{table}
\par
In \textbf{stage-2}, we report disentanglement and downstream prediction in Table~\ref{tab:quant_main} using an 8D VAE latent code for ADNI dataset. We supervise $\boldsymbol{z_{i,0}^{\,v}}$ with pseudo disease labels and $\boldsymbol{z_{i,1}^{\,v}}$ with available ages for our model. For $\beta$-VAE/$\beta$-TCVAE/DIP-VAE \cite{Locatello2019Challenging}, we report disease metrics on the latent dimension with the strongest disease association and keep age supervision fixed to $\boldsymbol{z_{i,1}^{\,v}}$. For PCA/ICA and HLLE+ICA \cite{horan2021when}, we do the same as for the VAEs, although it is not possible to embed age labels there. Our method improves disease and age separation (SAP) while maintaining reconstruction quality (CD) comparable to other VAEs. We report Pearson correlation, disease accuracy, and age RMSE from the designated disease and age latent dimensions using original labels, with a simple kNN \cite{knn} classifier and regressor evaluated on train and test splits. Ablations show that an adaptive SNNL temperature improves SAP over a fixed temperature and that covariance regularization improves factor separation. The last row reports OAI results for our method only, with percentages indicating improvement over the second-best model or reduction relative to the best model. Our model performs strongly on all metrics except reconstruction, which is expected under additional latent-space constraints \cite{Rabbi2024MELBA}.

\begin{figure}[t]
  \centering
  \includegraphics[height=7.5cm,width=11cm]{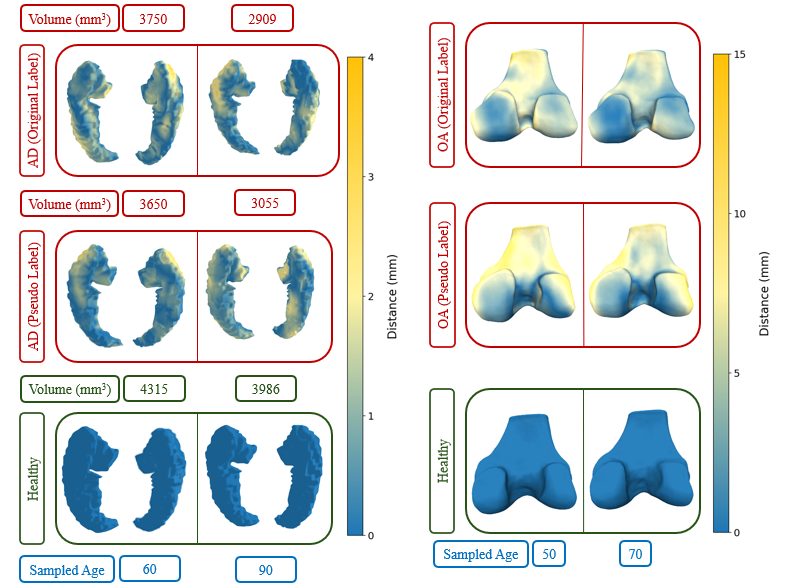}
  \caption{Rendering healthy and diseased shapes at different ages by latent traversal (trained by both real pseudo labels) shows consistent volume changes in ADNI (left) and deformation in OAI (right) that reflect the original dataset.}
  \label{fig:age_traversal}
\end{figure}

\subsection{Shape Generation through Latent Traversal and Label Mixing}
Fig.~\ref{fig:age_traversal} illustrate the model’s generative control. We sample multiple values along the age axis ($\boldsymbol{z_{i,1}^{\,v}}$) and, for each fixed $\boldsymbol{z_{i,1}^{\,v}}$, generate two shapes by setting $\boldsymbol{z_{i,0}^{\,v}}$ to its minimum and maximum values. This yields paired AD/CN shapes at matched ages, demonstrating controllable synthesis along both factors. Qualitatively, the generated pairs show volume ($\sim $30\% shrinkage due to AD) and disease-associated morphological differences aligning with known anatomical progression patterns reported in prior clinical shape studies \cite{franko2013evaluating}. Our self-supervised model closely follows the supervised model in terms of volume and shape changes \cite{franko2013evaluating}. In OAI, distal femur shape show limited change over time \cite{wise2020patterns}, which is consistent with our observation that age-related femoral deformation is not clearly separated in this cohort.
\par
We vary the fraction of real disease labels in Stage-2 (Table~\ref{tab:sap_label_mix}) for the ADNI dataset. Compared to leaving the remaining samples unlabeled, pseudo labels provide the largest benefit in low-label settings (0--30\% real labels: SAP 0.29--0.32 vs.\ 0.17--0.21). As real supervision increases ($\sim$80\%), the gain from pseudo-labels diminishes, and the results are reported on the test split. Here, our model supervised with 100\% real labels gives the upper-bound as we improve supervised disentanglement by incorporating recent methods.

\begin{table}[t]
\centering
\caption{Disease SAP across varying fractions of real AD labels used in Stage-2 training (first row). In the Real+No Label setting, the remaining samples are unlabeled, whereas in the Real+Pseudo setting, they are assigned pseudo-labels.}
\label{tab:sap_label_mix}
\setlength{\tabcolsep}{3.0pt}
\renewcommand{\arraystretch}{1.05}
\scriptsize
\begin{tabular*}{\linewidth}{@{\extracolsep{\fill}}lccccccccccc}
\hline
Real labels (\%) & 0 & 10 & 20 & 30 & 40 & 50 & 60 & 70 & 80 & 90 & 100 \\
\hline
SAP (Real + No label)   & - & 0.19 & 0.21 & 0.21  & 0.25 & 0.29 & 0.32 & 0.35 & \textbf{0.38} & \textbf{0.39} & \textbf{0.40}\\
SAP (Real + Pseudo) & \textbf{0.29} & \textbf{0.29} & \textbf{0.30} & \textbf{0.32}  & \textbf{0.32} & \textbf{0.34} & \textbf{0.35} & \textbf{0.37} & 0.37 & 0.38 & -\\
\hline
\end{tabular*}
\end{table}
\section{Conclusion}
We propose a two-stage, label-efficient framework for 3D shape disentanglement with limited or no diagnosis labels. Stage-1 INR learns clusterable implicit codes and yields pseudo disease labels through two-cluster discovery, and a stage-2 VAE separates disease and age into fixed latent dimensions while freezing the INR decoder as a renderer to preserve reconstruction. On ADNI hippocampus and OAI distal femur shapes, we outperform unsupervised baselines, enable controllable healthy and diseased shape generation across ages, and show pseudo labels help most in low-label settings but fade as real supervision increases in ADNI. In future work, we will study different diseases across diverse anatomical shapes.
%
%
%
\bibliographystyle{splncs04}
\bibliography{refs}

@article{Petersen2010ADNI,
  title={Alzheimer's disease Neuroimaging Initiative (ADNI) clinical characterization},
  author={Petersen, Ronald Carl and Aisen, Paul S and Beckett, Laurel A and Donohue, Michael C and Gamst, Anthony Collins and Harvey, Danielle J and Jack Jr, Clifford R and Jagust, William J and Shaw, Leslie M and Toga, Arthur W and others},
  journal={Neurology},
  volume={74},
  number={3},
  pages={201--209},
  year={2010},
  publisher={Lippincott Williams \& Wilkins}
}

@article{ambellan2019automated,
  title={Automated segmentation of knee bone and cartilage combining statistical shape knowledge and convolutional neural networks: Data from the Osteoarthritis Initiative},
  author={Ambellan, Felix and Tack, Alexander and Ehlke, Moritz and Zachow, Stefan},
  journal={Medical image analysis},
  volume={52},
  pages={109--118},
  year={2019},
  publisher={Elsevier}
}

@inproceedings{Mescheder2019Occupancy,
  title={Occupancy networks: Learning 3d reconstruction in function space},
  author={Mescheder, Lars and Oechsle, Michael and Niemeyer, Michael and Nowozin, Sebastian and Geiger, Andreas},
  booktitle={Proceedings of the IEEE/CVF conference on computer vision and pattern recognition},
  pages={4460--4470},
  year={2019}
}

@article{gropp2020igr,
  title={Implicit geometric regularization for learning shapes},
  author={Gropp, Amos and Yariv, Lior and Haim, Niv and Atzmon, Matan and Lipman, Yaron},
  journal={arXiv preprint arXiv:2002.10099},
  year={2020}
}

@article{burgess2018understanding,
  title={Understanding disentangling in $\beta $-VAE},
  author={Burgess, Christopher P and Higgins, Irina and Pal, Arka and Matthey, Loic and Watters, Nick and Desjardins, Guillaume and Lerchner, Alexander},
  journal={arXiv preprint arXiv:1804.03599},
  year={2018}
}

@article{chen2018isolating,
  title={Isolating sources of disentanglement in variational autoencoders},
  author={Chen, Ricky TQ and Li, Xuechen and Grosse, Roger B and Duvenaud, David K},
  journal={Advances in neural information processing systems},
  volume={31},
  year={2018}
}

@article{kumar2017variational,
  title={Variational inference of disentangled latent concepts from unlabeled observations},
  author={Kumar, Abhishek and Sattigeri, Prasanna and Balakrishnan, Avinash},
  journal={arXiv preprint arXiv:1711.00848},
  year={2017}
}

@inproceedings{Locatello2019Challenging,
  title={Challenging common assumptions in the unsupervised learning of disentangled representations},
  author={Locatello, Francesco and Bauer, Stefan and Lucic, Mario and Raetsch, Gunnar and Gelly, Sylvain and Sch{\"o}lkopf, Bernhard and Bachem, Olivier},
  booktitle={international conference on machine learning},
  pages={4114--4124},
  year={2019},
  organization={PMLR}
}

@article{Ouyang2022AgingDisease,
  title={Disentangling normal aging from severity of disease via weak supervision on longitudinal MRI},
  author={Ouyang, Jiahong and Zhao, Qingyu and Adeli, Ehsan and Zaharchuk, Greg and Pohl, Kilian M},
  journal={IEEE transactions on medical imaging},
  volume={41},
  number={10},
  pages={2558--2569},
  year={2022},
  publisher={IEEE}
}

@inproceedings{Kiechle2023ISBI,
  title={Explaining anatomical shape variability: supervised disentangling with a variational graph autoencoder},
  author={Kiechle, Johannes and Miller, Dylan and Slessor, Jordan and Pietrosanu, Matthew and Kong, Linglong and Beaulieu, Christian and Cobzas, Dana},
  booktitle={2023 IEEE 20th International Symposium on Biomedical Imaging (ISBI)},
  pages={1--5},
  year={2023},
  organization={IEEE}
}

@article{Rabbi2024MELBA,
  title={Disentangling hippocampal shape variations: A study of neurological disorders using mesh variational autoencoder with contrastive learning},
  author={Rabbi, Jakaria and Kiechle, Johannes and Beaulieu, Christian and Ray, Nilanjan and Cobzas, Dana},
  journal={arXiv preprint arXiv:2404.00785},
  year={2024}
}

@inproceedings{park2019deepsdf,
  title={Deepsdf: Learning continuous signed distance functions for shape representation},
  author={Park, Jeong Joon and Florence, Peter and Straub, Julian and Newcombe, Richard and Lovegrove, Steven},
  booktitle={Proceedings of the IEEE/CVF conference on computer vision and pattern recognition},
  pages={165--174},
  year={2019}
}

@article{horan2021when,
  title={When is unsupervised disentanglement possible?},
  author={Horan, Daniella and Richardson, Eitan and Weiss, Yair},
  journal={Advances in Neural Information Processing Systems},
  volume={34},
  pages={5150--5161},
  year={2021}
}

@article{bardes2022vicreg,
  title={Vicreg: Variance-invariance-covariance regularization for self-supervised learning},
  author={Bardes, Adrien and Ponce, Jean and LeCun, Yann},
  journal={arXiv preprint arXiv:2105.04906},
  year={2021}
}

@inproceedings{lucas2019posteriorcollapse,
  title={Understanding posterior collapse in generative latent variable models; 2019},
  author={Lucas, James and Tucker, George and Grosse, Roger and Norouzi, Mohammad},
  booktitle={URL https://openreview. net/forum},
  year={2019}
}

@incollection{lorensen1998marching,
  title={Marching cubes: A high resolution 3D surface construction algorithm},
  author={Lorensen, William E and Cline, Harvey E},
  booktitle={Seminal graphics: pioneering efforts that shaped the field},
  pages={347--353},
  year={1998}
}

@article{franko2013evaluating,
  title={Evaluating Alzheimer’s disease progression using rate of regional hippocampal atrophy},
  author={Frank{\'o}, Edit and Joly, Olivier and Alzheimer’s Disease Neuroimaging Initiative},
  journal={PloS one},
  volume={8},
  number={8},
  pages={e71354},
  year={2013},
  publisher={Public Library of Science San Francisco, USA}
}

@article{wise2020patterns,
  title={Patterns of change over time in knee bone shape are associated with sex},
  author={Wise, Barton L and Niu, Jingbo and Zhang, Yuqing and Liu, Felix and Pang, Joyce and Lynch, John A and Lane, Nancy E},
  journal={Clinical Orthopaedics and Related Research{\textregistered}},
  volume={478},
  number={7},
  pages={1491--1502},
  year={2020},
  publisher={LWW}
}

@article{cerrolaza2019computational,
  title={Computational anatomy for multi-organ analysis in medical imaging: A review},
  author={Cerrolaza, Juan J and Picazo, Mirella L{\'o}pez and Humbert, Ludovic and Sato, Yoshinobu and Rueckert, Daniel and Ballester, Miguel {\'A}ngel Gonz{\'a}lez and Linguraru, Marius George},
  journal={Medical image analysis},
  volume={56},
  pages={44--67},
  year={2019},
  publisher={Elsevier}
}

@article{gerardin2009multidimensional,
  title={Multidimensional classification of hippocampal shape features discriminates Alzheimer's disease and mild cognitive impairment from normal aging},
  author={Gerardin, Emilie and Ch{\'e}telat, Ga{\"e}l and Chupin, Marie and Cuingnet, R{\'e}mi and Desgranges, B{\'e}atrice and Kim, Ho-Sung and Niethammer, Marc and Dubois, Bruno and Leh{\'e}ricy, St{\'e}phane and Garnero, Line and others},
  journal={Neuroimage},
  volume={47},
  number={4},
  pages={1476--1486},
  year={2009},
  publisher={Elsevier}
}

@inproceedings{dannecker2024cina,
  title={Cina: Conditional implicit neural atlas for spatio-temporal representation of fetal brains},
  author={Dannecker, Maik and Kyriakopoulou, Vanessa and Cordero-Grande, Lucilio and Price, Anthony N and Hajnal, Joseph V and Rueckert, Daniel},
  booktitle={International Conference on Medical Image Computing and Computer-Assisted Intervention},
  pages={181--191},
  year={2024},
  organization={Springer}
}

@inproceedings{molaei2023implicit,
  title={Implicit neural representation in medical imaging: A comparative survey},
  author={Molaei, Amirali and Aminimehr, Amirhossein and Tavakoli, Armin and Kazerouni, Amirhossein and Azad, Bobby and Azad, Reza and Merhof, Dorit},
  booktitle={Proceedings of the IEEE/CVF International Conference on Computer Vision},
  pages={2381--2391},
  year={2023}
}

@inproceedings{zhang2025surface,
  title={Surface-Based Multi-axis Longitudinal Disentanglement Using Contrastive Learning for Alzheimer’s Disease},
  author={Zhang, Jianwei and Shi, Yonggang},
  booktitle={International Conference on Medical Image Computing and Computer-Assisted Intervention},
  pages={585--594},
  year={2025},
  organization={Springer}
}

@article{amiranashvili2024learning,
  title={Learning continuous shape priors from sparse data with neural implicit functions},
  author={Amiranashvili, Tamaz and L{\"u}dke, David and Li, Hongwei Bran and Zachow, Stefan and Menze, Bjoern H},
  journal={Medical image analysis},
  volume={94},
  pages={103099},
  year={2024},
  publisher={Elsevier}
}

@inproceedings{lee2022regularized,
  title={Regularized autoencoders for isometric representation learning},
  author={Lee, Yonghyeon and Yoon, Sangwoong and Son, Minjun and Park, Frank C},
  booktitle={International Conference on Learning Representations},
  year={2022}
}

@inproceedings{akiba2019optuna,
  title={Optuna: A next-generation hyperparameter optimization framework},
  author={Akiba, Takuya and Sano, Shotaro and Yanase, Toshihiko and Ohta, Takeru and Koyama, Masanori},
  booktitle={Proceedings of the 25th ACM SIGKDD international conference on knowledge discovery \& data mining},
  pages={2623--2631},
  year={2019}
}

@article{knn,
  title={Application of k-nearest neighbor (knn) approach for predicting economic events: Theoretical background},
  author={Imandoust, Sadegh Bafandeh and Bolandraftar, Mohammad and others},
  journal={International journal of engineering research and applications},
  volume={3},
  number={5},
  pages={605--610},
  year={2013}
}

@article{chupin2009fully,
  title={Fully automatic hippocampus segmentation and classification in Alzheimer's disease and mild cognitive impairment applied on data from ADNI},
  author={Chupin, Marie and G{\'e}rardin, Emilie and Cuingnet, R{\'e}mi and Boutet, Claire and Lemieux, Louis and Leh{\'e}ricy, St{\'e}phane and Benali, Habib and Garnero, Line and Colliot, Olivier},
  journal={Hippocampus},
  volume={19},
  number={6},
  pages={579--587},
  year={2009},
  publisher={Wiley Online Library}
}
%

\end{document}